\documentclass{article}

\PassOptionsToPackage{numbers,sort&compress}{natbib}

 \usepackage[dblblindworkshop,final]{neurips_2025}
\workshoptitle{2nd Workshop on Multi-modal Foundation Models and Large Language Models for Life Sciences}

\usepackage[utf8]{inputenc} 
\usepackage[T1]{fontenc}    
\usepackage{hyperref}       
\usepackage{url}            
\usepackage{booktabs}       
\usepackage{amsfonts}       
\usepackage{nicefrac}       
\usepackage{microtype}      
\usepackage{xcolor}         
\usepackage{graphicx}
\usepackage{amsmath}
\usepackage{dsfont}
\usepackage{diagbox}

\title{Uncertainty-Guided Model Selection for Tabular Foundation Models in Biomolecule Efficacy Prediction}

\author{%
  Jie Li \\
  AIML\\
  GSK\\
  South San Francisco, CA, USA 94080 \\
  \texttt{jerry.8.li@gsk.com} \\
  \And
  Andrew McCarthy \\
  Molecular Modalities Discovery \\
  GSK\\
  Cambridge, MA, USA 02140 \\
  \texttt{andrew.p.mccarthy@gsk.com} \\
  \AND
  Zhizhuo Zhang \\
  AIML \\
  GSK\\
  South San Francisco, CA, USA 94080 \\
  \texttt{zhizhuo.x.zhang@gsk.com} \\
  \And
  Stephen Young \\
  AIML \\
  GSK\\
  London, UK, N1C 4AG \\
  \texttt{stephen.r.young@gsk.com} \\
}

\begin{document}

\maketitle

\begin{abstract}
In-context learners like TabPFN are promising for biomolecule efficacy prediction, where established molecular feature sets and relevant experimental results can serve as powerful contextual examples. However, their performance is highly sensitive to the provided context, making strategies like post-hoc ensembling of models trained on different data subsets a viable approach. An open question is how to select the best models for the ensemble without access to ground truth labels. In this study, we investigate an uncertainty-guided strategy for model selection. We demonstrate on an siRNA knockdown efficacy task that a TabPFN model using straightforward sequence-based features can surpass specialized state-of-the-art predictors. We also show that the model's predicted inter-quantile range (IQR), a measure of its uncertainty, has a negative correlation with true prediction error. We developed the OligoICP method, which selects and averages an ensemble of models with the lowest mean IQR for siRNA efficacy prediction, achieving superior performance compared to naive ensembling or using a single model trained on all available data. This finding highlights model uncertainty as a powerful, label-free heuristic for optimizing biomolecule efficacy predictions.
\end{abstract}

\section{Introduction}
Predicting biomolecule efficacy is a central challenge in drug discovery, with the potential to significantly accelerate pharmaceutical development by prioritizing promising candidates before wet-lab validation. However, biomolecule efficacy datasets are often small, heterogeneous, and collected using varied experimental techniques\cite{bio_challenges}. Such inconsistency complicates the training of reliable predictive models, hindering the real-world influence of these models.

While domain-specific in-context learners (ICLs) have shown promise for few-shot molecular property prediction\cite{icl_molecular}, recent general-purpose tabular models like TabPFN\cite{tabpfn}, TabPFNv2\cite{tabpfnv2} and TabDPT\cite{tabdpt} offer a new approach that can enhance biomolecule efficacy prediction by leveraging relevant data as context. However, choosing the right data is not always straightforward. Simply using more data does not guarantee better performance, and large datasets can exceed an ICL's practical limits. For example, TabPFN's computational cost scales quadratically with the number of training examples, making it infeasible to process very large training sets\cite{tabicl}. Furthermore, using a context size that is out-of-distribution from the model's pre-training data can also lead to inferior results.

Strategies to address this include data retrieval and fine-tuning, as exemplified by LoCalPFN \cite{localpfn}, or post-hoc ensembling (PHE)\cite{post_hoc_ensembling}, which combines predictions from models trained on different data subsets. However, effective ensembling still requires a reliable method for selecting best performing models. In this paper, we introduce OligoICP (Oligo In-Context Predictor), an uncertainty based strategy for model selection in PHE for siRNA knockdown efficacy prediction. The core principle of OligoICP is that a model's predicted uncertainty can serve as a proxy for its true prediction error, allowing us to select and combine models with the lowest predicted uncertainty to form a final, more accurate ensemble. While prior work has used uncertainty of sensitive attributes to guide sample selection to improve fairness in ICLs \cite{fairicl}, its use for improving model accuracy by model selection has not been explored to the best of our knowledge. We demonstrated that our method, based on TabPFN model with straightforward features, surpasses specialized, state-of-the-art models. Furthermore, we show the model’s inter-quantile range (IQR) reflects its prediction uncertainty and negatively correlates with true prediction error, making it a competitive metric for selecting models that yield optimal performance.

\section{Dataset and Features}
Small interfering RNAs (siRNAs) are a promising therapeutic modality that silences target genes by cleaving mRNA transcripts\cite{sirna}. A key challenge is designing siRNAs with high knockdown efficacy. We used the dataset composed by Huesken et al.\cite{huesken}, which has been used to develop multiple siRNA efficacy predictors, including the state of the art model OligoFormer\cite{oligoformer}. We divided the Huesken dataset (2361 data points) into 29 subsets, each corresponding to a different mRNA target. Additionally, we collected data on a novel target not present in the Huesken dataset (Target1) from public accessible patents from three institutions denoted as institution A, B and C, with 295, 366 and 9 data points, respectively. We also collected 252 data points from a patent published by institution D on a second novel target (Target2).

Following the protocol from OligoFormer, we represented siRNA molecules as 19-mers, and prepared the corresponding mRNA centered around the region that the siRNA reverse complements to, with a flanking region of 19 nucleotides on both sides. Our feature set includes: (1) one-hot representation for each nucleotide in the siRNA and mRNA sequences, (2) counts of all possible ribonucleotide trimers from both sequences, and (3) thermodynamic parameters that describe the siRNA stability and binding affinity of the siRNA-mRNA interactions. A detailed description of the feature formulation can be found in Appendix A.  This resulted in 574 features per data point. Although the number of features exceeds TabPFN’s pre-training limit, we found that using the full feature set with the \verb|"ignore_pretraining_limits=True"| flag yielded higher performance than feature subsetting.

\section{Experiment and Results}
\subsection{Comparing TabPFN Model with a State-of-the-Art Model}
We validated the use of TabPFN with custom features by comparing its performance to OligoFormer\cite{oligoformer}, a state-of-the-art model for siRNA knockdown efficacy prediction. To assess in-distribution performance, we evaluated the mean absolute error (MAE) and Pearson correlation coefficient on the Huesken dataset using a 5-fold cross-validation approach for TabPFN. As shown in Table \ref{tab:sota}, the cross-validated TabPFN model surpasses the performance of OligoFormer evaluated on its own training set, supporting the use of TabPFN for this task.

Given that these models are usually used in out-of-distribution settings, we then explored the capability for the TabPFN model to generalize with few-shot data on another target (Target1) that is distinct from those in the Huesken dataset. We used the smallest context (subset C) with only 9 data points to evaluate prediction MAEs and correlation coefficients for the two larger subsets (A and B). The results are also provided in Table \ref{tab:sota}. Despite the lower performance compared with the Huesken dataset, it still does significantly better than OligoFormer. This highlights the real-world applicability of TabPFN and the potential limitations of specialized models like OligoFormer when faced with novel targets.

\begin{table}[h!]
  \caption{Mean absolute errors (MAEs) and correlation coefficients for TabPFN \cite{tabpfnv2} and OligoFormer \cite{oligoformer}. In-distribution performance (Huesken) for TabPFN is reported as mean $\pm$ 95\% confidence interval based on 5-fold cross validation. Out-of-distribution (Target1) performance for TabPFN is based on few-shot prediction using the same target as context. In both cases, TabPFN outperforms the specialized SOTA model OligoFormer.}
  \label{tab:sota}
  \centering
  \begin{tabular}{lllll}
    \toprule
   &\multicolumn{2}{c}{\textbf{MAE ($\downarrow$)} }&\multicolumn{2}{c}{\textbf{Corr. Coef ($\uparrow$)}}\\
    Dataset&TabPFN&OligoFormer&TabPFN&OligoFormer \\
    \midrule
    Huesken & \textbf{0.087 $\pm $0.004} & 0.096 & \textbf{0.677$\pm$ 0.042} & 0.630\\
    \midrule
    Target1 (A) & \textbf{0.245}  & 0.251   &\textbf{0.244}&  0.158\\
    Target1 (B)    & \textbf{0.159} & 0.180 &\textbf{0.200}&    0.082\\

    \bottomrule
  \end{tabular}
\end{table}

\subsection{Using Model Inter-Quantile Range for Prediction Quality Estimation}
A TabPFN regressor can output not just a point estimate, but it can generate a distribution of possible values and provide lower bound and upper bound estimations for a certain quantile level\cite{tabpfn}. We confirmed that our model is well-calibrated (see Appendix B), meaning its quantile estimates accurately reflect the probability of the true value falling within a given range. In this study, we use the 15\% and 85\% quantiles, expecting a 70\% chance of correctness when the data is in-distribution. The inter-quantile range (IQR) is then defined as the difference between 85\% quantile and 15\% quantile from model predictions. The IQR serves as a measure of the model's intrinsic uncertainty; we investigated whether it could be used to approximate the true prediction error.

First, we combined all available data, used a randomly selected 70\% of data for training, and predicted efficacy values for the rest 30\% of data using the TabPFN model. The MAEs are plotted against IQRs in Figure 1a for the test data. We could see a clear trend between the IQR and MAE, especially in the middle IQR range. A higher IQR leads to the distribution of MAE towards higher values, which means the predictions are less accurate. Because there is little data falling into the smallest and largest IQR data ranges, the estimates are noisy and inconclusive of the trend in these ranges. We would like to point out a single data point with lower IQR does not necessarily mean it will have high prediction accuracy, but rather it has a higher chance to be accurate compared to data with higher IQR.

We then explored whether the IQRs can be used to differentiate models with higher and lower accuracies. Using Target1 (A) as an example, we generated eight different random permutations of training subset orders, and trained a collection of models by progressively adding more training subsets as context following the predefined random order, until every training subset was included. Figure 1b plots the correlation coefficient of each model against the mean IQR of its predictions. An observable negative correlation (Pearson's $r=-0.42$) emerges, confirming that models exhibiting lower overall uncertainty (lower mean IQR) tend to produce more accurate predictions. Crucially, this IQR metric is calculated without knowledge of the ground truth labels, making it a viable heuristic for model selection.

\begin{figure}[h!]
  \centering
  \includegraphics[width=0.98\textwidth]{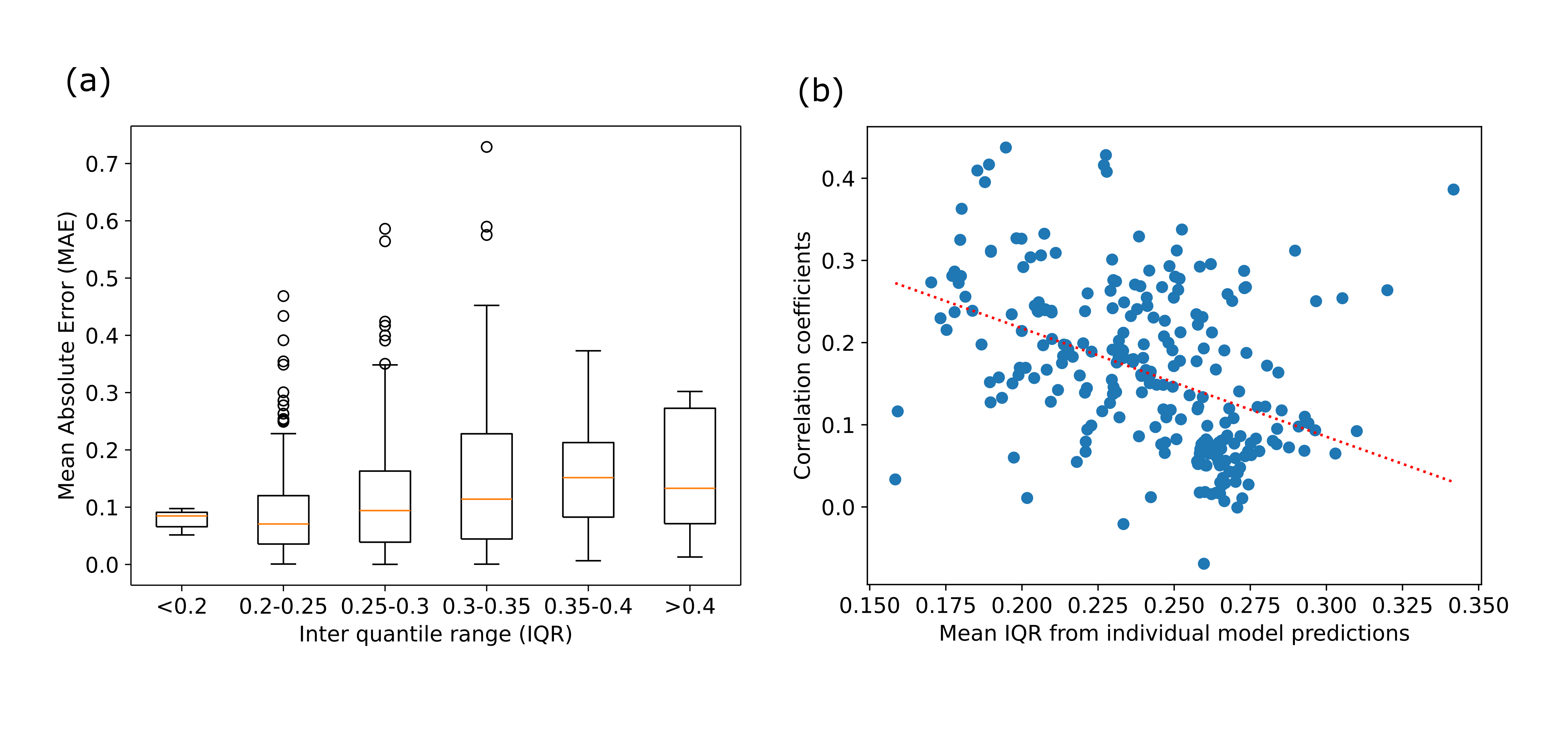}
  \caption{\textit{Using inter-quantile ranges (IQRs) to evaluate prediction quality.} (a) Distribution of mean absolute error (MAE) categorized by (IQR) on randomly chosen held-out test set.  Higher IQR is associated with higher error. (b) Scatter plot showing the negative correlation between a model's performance (correlation coefficient) and its mean prediction IQR on the Target1 (A) dataset. The dotted line shows the best linear fit ($r=-0.42$).}
  
\end{figure}

\subsection{Model Selection with Inter-Quantile Range}
Post-hoc ensembling (PHE) is a common technique to improve model performance by averaging predictions from multiple models trained with different model hyperparameters and different datasets. Given that IQR is indicative of model performance, we developed the OligoICP framework, which uses IQR as a metric for model selection. For each test set, we trained an ensemble of 400 models, where each model was trained on $k$ randomly selected training subsets ($k$ was chosen randomly between 1 and 20 for each model). The process was repeated 3 times using different random seeds. We then compared three strategies: (1) a single baseline model trained on all available training data (\textbf{All data single model}), (2) a naive ensemble that averages predictions from all 400 models (\textbf{Full ensemble mean}), and (3) OligoICP, which averages predictions from only the top 10\% of models with the lowest mean IQR (\textbf{Top 10\% ensemble mean}). All results are given as means and 95\% confidence intervals of 3 parallel runs.
\begin{table}[h!]
  \caption{Mean absolute errors (MAEs) for different datasets with different model selection strategies. For the ensemble strategies, results are reported as means and 95\% confidence intervals of 3 individual runs using different random seeds. }
  \label{tab:mae}
  \centering
  \begin{tabular}{lllll}
    \toprule
    \textbf{Dataset}&\textbf{OligoICP (Top 10\% } &\textbf{Full ensemble }&\textbf{All data }&\textbf{Ensemble best} \\
    &\textbf{ensemble mean)}&\textbf{mean}&\textbf{single model}&(oracle result)\\
    \midrule
    Target1 (A) & 0.270 $\pm$ 0.005  & 0.268 $\pm$ 0.002   &0.278&  0.197\\
    Target1 (B)& 0.174 $\pm$ 0.001 & 0.169 $\pm$ 0.001 &0.172&    0.149\\
    Target2&0.185 $\pm$ 0.001&0.189 $\pm$ 0.001&0.186&0.161\\

    \bottomrule
  \end{tabular}
\end{table}
\begin{table}[h!]
  \caption{Correlation coefficients for different datasets with different model selection strategies. For the ensemble strategies, results are reported as means and 95\% confidence intervals  of 3 individual runs using different random seeds.}
  \label{tab:corr}
  \centering
  \begin{tabular}{lllll}
    \toprule
    \textbf{Dataset}&\textbf{OligoICP (Top 10\% } &\textbf{Full ensemble }&\textbf{All data }&\textbf{Ensemble best} \\
    &\textbf{ensemble mean)}&\textbf{mean}&\textbf{single model}&(oracle result)\\
    \midrule
    Target1 (A) & 0.278 $\pm$ 0.015  & 0.257 $\pm$ 0.012   &0.051&  0.544\\
    Target1 (B)& 0.072 $\pm$ 0.005 & 0.086 $\pm$ 0.020 &0.112&    0.430\\
    Target2&0.246 $\pm$ 0.015&0.230 $\pm$ 0.002&0.230&0.384\\

    \bottomrule
  \end{tabular}
\end{table}

The results are summarized in Table \ref{tab:mae} (MAE) and Table \ref{tab:corr} (Correlation). While the OligoICP approach has a marginal effect on MAE, it provides a notable improvement in correlation for two of the three test datasets. For Target1 (A), OligoICP yields a correlation of 0.278 $\pm$ 0.015, a substantial improvement over the performance of a single model trained with all data (correlation of 0.051). Similarly, for Target2, OligoICP achieves the highest correlation, highlighting the effectiveness of our ensemble selection approach. While Target1 (B) remains a challenging case for all methods, these results demonstrate that IQR-based model selection is a promising strategy. We would like to emphasize that the extra amount of compute is acceptable, given that each model in the ensemble only operates on a limited size of data, and the inference with different models in the ensemble is embarrassingly parallelizable. Furthermore, OligoICP provides a direct way to handle large amount of context data that do not fit into a single forward pass of the TabPFN model. The "\textbf{Ensemble Best}" column shows the performance of the single best model in the ensemble from all 3 parallel runs. It is an oracle result which means there is no way to identify a single best model \textit{a priori} for new test data. Comparing our approach with the "ensemble best" reveals the upper bound of performance and the remaining gap for improvement in model selection strategies.

\section{Related Works}

\textbf{Tabular In-Context Learning}
The development of tabular foundation models began with TabPFN \cite{tabpfn}, a Transformer architecture pre-trained on synthetic data that established the viability of in-context learning for small tabular problems using deep learning models. This was followed by improvements in pre-training, such as using real-world data in TabDPT \cite{tabdpt} for better generalization, and architectural refinements in TabPFN v2 \cite{tabpfnv2}. A key line of subsequent research has focused on overcoming the scalability limitations imposed by the Transformer's quadratic complexity. Proposed solutions include retrieval and fine-tuning on local data subsets, as in LoCalPFN \cite{localpfn}, and new architectures designed for larger contexts like TabICL \cite{tabicl}. Our work aligns with a more application focused research direction which adapts these general models for a specialized scientific domain.

\textbf{Model Selection for Post-Hoc Ensembling}
Our work on uncertainty-guided model selection builds upon a rich history of post-hoc ensembling techniques. Foundational approaches include constructing ensembles from models saved during hyperparameter optimization, as demonstrated in Auto-sklearn \cite{post_hoc_ensembling}, and dynamic weighted regressors that adjust model contributions based on performance metrics like RRMSE \cite{dwr, RRMSE_voting}.

More recently, sophisticated methods for model re-weighting and pruning have emerged. PSEO (Post-hoc Stacking Ensemble Optimization) frames model selection as a Bayesian hyperparameter optimization problem \cite{pseo}, while other work has employed end-to-end neural networks to dynamically re-weight base models \cite{neural_ensembler}. Our method is particularly related to approaches that leverage model uncertainty. For instance, \cite{dynamic_ensemble_selection} used network uncertainty estimations to select a dynamic ensemble for improved adversarial robustness. Conformal prediction, which uses a small calibration set to provide statistically rigorous uncertainty bounds, has also been explored for model selection and aggregation \cite{conformal_1, conformal_2, conformal_3}.

\textbf{siRNA Knockdown Efficacy Prediction}
Computational prediction of siRNA efficacy has been a long-standing goal to accelerate therapeutic development. Early efforts utilized classical machine learning models \cite{iscore, DMIR,svm,random_forest}, and more recently a variety of neural architectures have been applied \cite{huesken,sidpt,deepsipred,deepsilencer,oligoformer}. Currently, a state-of-the-art predictor is OligoFormer \cite{oligoformer}, which integrates thermodynamic features, pre-trained RNA-FM embeddings \cite{rnafm}, and a custom Transformer-based encoder. Our work demonstrates that a general-purpose tabular foundation model can achieve competitive, and in some cases superior, performance compared to this specialized model.

\section{Conclusion}
We have shown that in-context learners like TabPFN, equipped with straightforward engineered features, could surpass specialized state-of-the-art models on siRNA knockdown efficacy prediction. Our primary contribution is the development and validation of OligoICP, a method that uses the model's self-reported uncertainty, quantified by the inter-quantile range (IQR) of its predictions as a potent, label-free heuristic for post-hoc model selection. We showed that the OligoICP ensemble,  composed of models selected for their low prediction uncertainty, can achieve superior performance, particularly in terms of correlation, compared to a single model using all available data or a naive ensemble average. This approach offers a practical strategy to enhance the reliability of in-context learners in real-world scenarios where ground truth labels are unavailable for model tuning. Future work should aim to narrow the gap to the oracle "ensemble best" by further iterating on the algorithm, and by validating this uncertainty-guided methodology across a broader range of biomolecule prediction tasks.

\bibliographystyle{unsrtnat}
\bibliography{references}
\newpage
\appendix
\section{Details about data preparation and features used for TabPFN model}
Small interfering RNA molecules (siRNAs) naturally exist as double stranded oligonucleotides, and we take the antisense strand (the strand that reverse complements to the mRNA transcript) as our input to the model. We ensure that siRNA molecules are 19 nucleotides (nts) in length. For any molecule longer than 19 nts from the patents, we take the first 19 nts after the leading uracil (U) as U is typically not part of the siRNA that matches to mRNA. We remove any siRNA shorter than 19 nts. For the mRNA, we take a slice from the transcript centered around the region that siRNA binds to, plus flanking regions of 19 nts on both sides. This adds up to 57 nts for the mRNA sequence input. When the flanking region goes beyond the transcript, "X"s are added in the places where nts are missing. Figure 2 gives a schematic explanation of the siRNA and mRNA sequences used to generate features.

\begin{figure}[h!]
  \centering
  \includegraphics[width=0.8\textwidth]{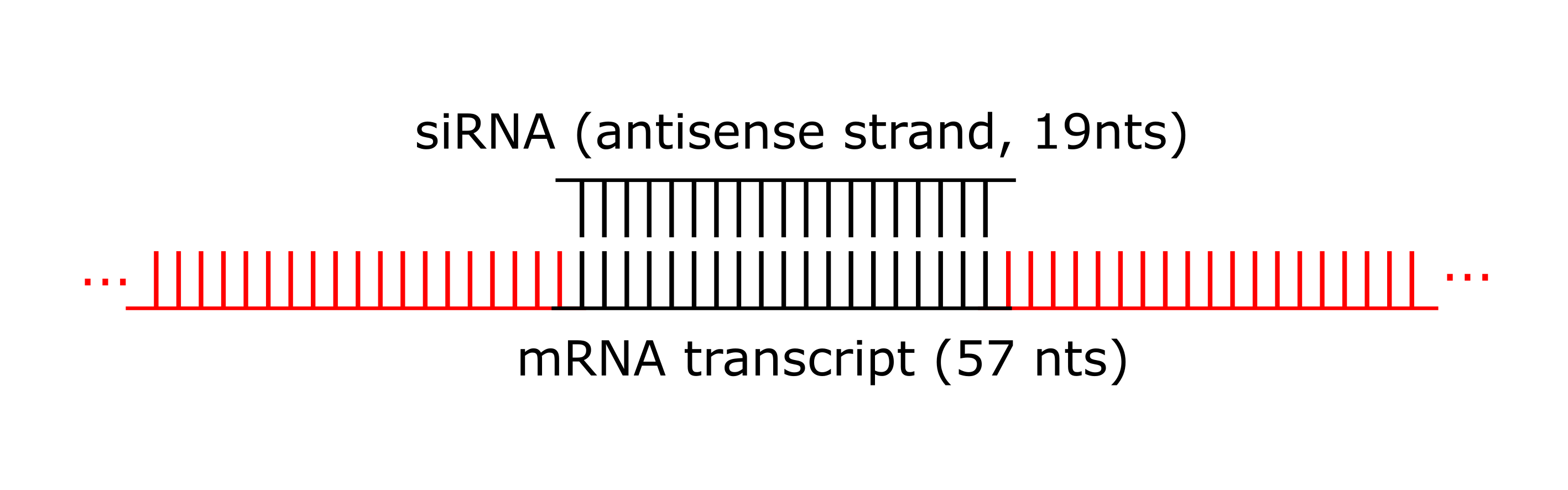}
  \caption{Schema of the siRNA and mRNA taken as input to the TabPFN model }
  
\end{figure}

We generated straightforward features for both the siRNA and mRNA sequences, which include one-hot features:
$$F_{\text{one-hot}}=\left( \bigoplus_{i=1}^{19} \text{OHE}_{\text{AUCG}}(S_{\text{siRNA}}[i]) \right)\oplus \left( \bigoplus_{j=1}^{57} \text{OHE}_{\text{AUCGX}}(S_{\text{mRNA}}[j]) \right)$$
where $S_{\text{siRNA}}[i]$ represents the $i$th position in the siRNA sequence, $S_{\text{mRNA}}[j]$ represents the $j$th position in the mRNA sequence, and $\text{OHE}(\mathbf{N})$ is the one-hot encoding for a nucleotide $\mathbf{N}\in\{\text{A, U, C, G}\}$ for siRNA and $\mathbf{N}\in\{\text{A, U, C, G, X}\}$ for mRNA. $\oplus$ represents feature concatenation. This corresponds to $4\times19+5\times57=361$ features.

We also include trimer count features for all possible trimers from the siRNA and mRNA vocabularies. Let $T_{\text{siRNA}}$ be the set of all $4^3=64$ trimer combinations for siRNA and $T_{\text{mRNA}}$ be the set of $5^3=125$ trimer combinations for mRNA, then the trimer count features are:
$$F_{\text{trimer}}=\left( \bigoplus_{t \in T_{\text{siRNA}}} \sum_{i=1}^{17}\mathds{1}(S_{\text{siRNA}}[i:i+2]= t) \right) \oplus \left( \bigoplus_{t \in T_{\text{mRNA}}}\sum_{j=1}^{55} \mathds{1}(S_{\text{mRNA}}[j:j+2]= t) \right)$$
where $S_{\text{siRNA}}[i:i+2]$ represents the three consecutive nucleotides in the siRNA sequence starting at position $i$ and $S_{\text{mRNA}}[j:j+2]$ represents the three consecutive nucleotides in the mRNA sequence starting at position $j$. $\mathds{1}(a=b)$ is 1 when $a=b$ evaluates to true, and 0 otherwise. This corresponds to $64+125=189$ features.

Finally, the thermodynamic parameters follows the same implementation as in OligoFormer\cite{oligoformer}, and are only calculated based on the siRNA sequence. These features include single and dinucleotide features:

\begin{align*}
\mathbf{N}(k) &= \begin{cases} 0, & S_{\text{siRNA}}[k] \neq \mathbf{N} \\ 1, & S_{\text{siRNA}}[k] = \mathbf{N} \end{cases} \quad \mathbf{N} \in \{\mathbf{A, U, C, G}\}, k \in [1,19] \\
\\
\mathbf{NM}(k) &= \begin{cases} 0, & S_{\text{siRNA}}[k:k+1] \neq \mathbf{NM} \\ 1, & S_{\text{siRNA}}[k:k+1] = \textbf{NM} \end{cases} \quad \textbf{N,M} \in \{\textbf{A, U, C, G}\}, k \in [1,18] \\
\\
\textbf{N}(all) &= \frac{\sum_{k=1}^{19}[\textbf{N}(k)]}{19}, \quad \textbf{N} \in \{A, U, C, G\} \\
\\
\textbf{NM}(all) &= \frac{\sum_{k=1}^{18}[\textbf{NM}(k)]}{18}, \quad \textbf{N,M} \in \{A, U, C, G\}
\end{align*}

A second part of the thermodynamic parameters are calculated from the Gibbs free energy changes ($\Delta G$) and enthalpy changes ($\Delta H$). These are defined for each dinucleotide as \cite{thermo}

\begin{table}[h!]
  \caption{$\Delta G$ values for dinucleotides used for thermodynamic parameters calculation ($kcal/mol$)}
  \centering
  \begin{tabular}{l|cccc}
    \toprule
   \diagbox{First nt}{Second nt}&A&U&C&G\\
   \midrule
    A&-0.93&-1.10&-2.24&-2.08 \\
    U&-1.33&-0.93&-2.35&-2.11\\
    C&-2.11&-2.08&-3.26&-2.36\\
    G&-2.35&-2.24&-3.42&-3.26\\

    \bottomrule
  \end{tabular}
\end{table}

\begin{table}[h!]
  \caption{$\Delta H$ values for dinucleotides used for thermodynamic parameters calculation ($kcal/mol$)}
  \centering
  \begin{tabular}{l|cccc}
    \toprule
   \diagbox{First nt}{Second nt}&A&U&C&G\\
   \midrule
    A&-6.82&-9.38&-11.40&-10.48 \\
    U&-7.69&-6.82&-12.44&-10.44\\
    C&-10.44&-10.48&-13.39&-10.64\\
    G&-12.44&-11.40&-14.882&-13.39\\

    \bottomrule
  \end{tabular}
\end{table}

There are also three special thermodynamic parameters, calculated as
\begin{align*}
\Delta G_{all}&=\Delta G_{init}+\Delta G_{end}\times n_{\text{A/U end}}+\Delta G_{sym} + \sum_{i=1}^{18}\Delta G(S_{\text{siRNA}}[k:k+1])\\
\Delta H_{all}&=\Delta H_{init}+\Delta H_{end}\times n_{\text{A/U end}}+ \sum_{i=1}^{18}\Delta H(S_{\text{siRNA}}[k:k+1])\\
\Delta \Delta G_{all}&=\Delta G(S_{\text{siRNA}}[1:2]) -\Delta G(S_{\text{siRNA}}[18:19]) +\Delta G_{end}\times n_{\text{A/U end}}
\end{align*}
where $n_{\text{A/U end}}=\mathbf{A}(1)+\mathbf{U}(1)+\mathbf{A}(19)+\mathbf{U}(19)$ is the count of A and U nucleotides at the ends, $\Delta G_{init}=4.09$, $\Delta G_{end}=0.45$, $\Delta H_{init}=3.61$, $\Delta H_{end}=3.72$, and $\Delta G_{sym}=0.43$ (all in $kcal/mol$) if the siRNA sequence is its own reverse complement, and zero otherwise.

The full set of thermodynamic features consist of 
\begin{align*}
    F_{thermo}=[&\Delta \Delta G_{all},\Delta G(S_{\text{siRNA}}[1:2]),\Delta H(S_{\text{siRNA}}[1:2]),\mathbf{U}(1), \mathbf{G}(1), \\
    &\Delta H_{all},\mathbf{U}(all),\mathbf{UU}(1),\mathbf{G}(all), \mathbf{GG}(1), \mathbf{GC}(1),\mathbf{GG}(all),\\
    &\Delta G(S_{\text{siRNA}}[2:3]),\mathbf{UA}(all),\mathbf{U}(2), \mathbf{C}(1),\mathbf{CC}(all),\\
    &\Delta G(S_{\text{siRNA}}[18:19]), \mathbf{CC}(1),\mathbf{GC}(all),\mathbf{CG}(1),\\
    &\Delta G(S_{\text{siRNA}}[13:14]),\mathbf{UU}(all),\mathbf{A}(19)]
\end{align*}
which has a length of 24.

The final input feature is the combination of one-hot features, trimer count features and thermodynamic features:
$$F_{input}=F_{one-hot}\oplus F_{trimer}\oplus F_{thermo}$$
and the length is $361+189+24=574$.

\section{TabPFN Model Calibration with siRNA Data}
To show that the model is well calibrated with the siRNA data, we combined all data and performed a 70\%-30\% train-test split. We ran inference on the test data using the train data as context, obtaining lower and upper bound predictions at different quantile levels using \verb|reg.predict(test_X, output_type="quantiles", quantiles=[l, u])|, where \verb|reg| is the trained TabPFN model and $l, u$ are the lower and upper quantile bounds. We then calculated the empirical coverage as the fraction of true labels that fall within the predicted quantile range: $$\text{Coverage} = \frac{1}{N}\sum_{i=1}^{N}\mathds{1}(y_{\text{pred,lower}}^{(i)}<y_{\text{true}}^{(i)}<y_{\text{pred,upper}}^{(i)}).$$

\begin{figure}[h!]
  \centering
  \includegraphics[width=0.8\textwidth]{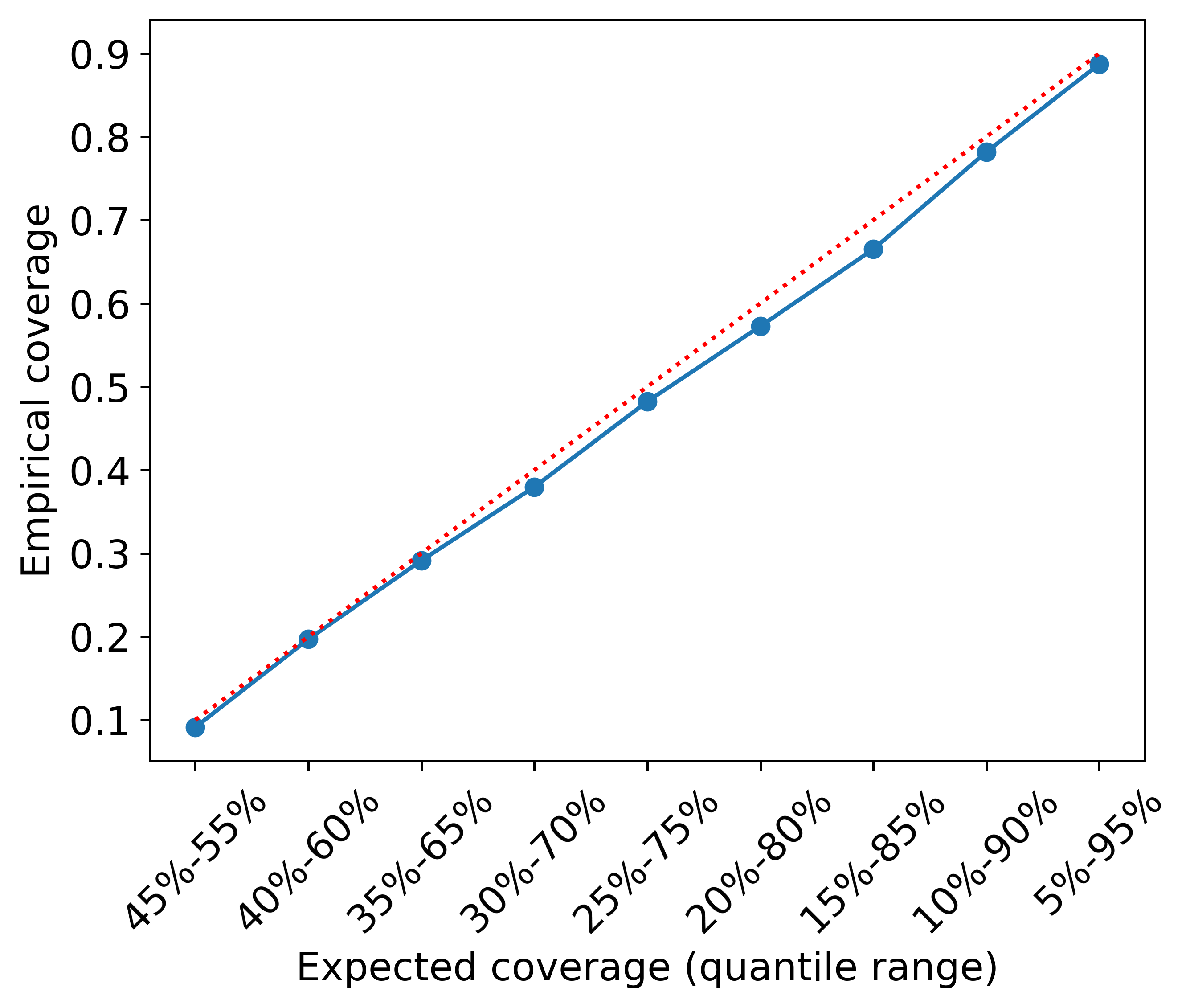}
  \caption{Relationship between empirical coverage and the expected quantile range provided to the model. The close adherence to the identity line ($y=x$) demonstrates that the model is well-calibrated.}
  \label{fig:calib}
\end{figure}

The empirical coverage was plotted against the expected coverage (the quantile range, $u-l$) in Figure \ref{fig:calib}. The curve closely follows the diagonal line, indicating the model is well-calibrated, meaning its uncertainty estimates are reliable.

\end{document}